\documentclass[letterpaper, 10 pt, conference]{ieeeconf}  % Comment this line out if you need a4paper

\IEEEoverridecommandlockouts                              % This command is only needed if 
\usepackage{graphics}    % for pdf, bitmapped graphics files
\usepackage{epsfig}      % for postscript graphics files
\usepackage{times}
\usepackage{amsmath}
\usepackage{amssymb}

\usepackage{cite}

\makeatletter
\let\NAT@parse\undefined
\makeatother

\usepackage{xcolor}
\usepackage{booktabs}
\usepackage{tabularx}
\usepackage{ulem} % for \sout
\usepackage{algorithm}
\usepackage{algpseudocode}
\usepackage{graphicx}
\usepackage[font=small,labelfont=bf]{caption}
\usepackage[inkscapelatex=false]{svg}
\usepackage[export]{adjustbox}
\usepackage{siunitx}
\let\labelindent\relax \usepackage{enumitem}
\usepackage{cuted}
\usepackage{capt-of}   % for \captionof
\usepackage{hyperref}
\hypersetup{
  colorlinks=true,
  citecolor=oopsiecolor,
  linkcolor=oopsiecolor,
  urlcolor=oopsiecolor
}

\newcommand{\ours}{\textsc{OopsieRecover}}

\usepackage{xcolor}
\usepackage{lipsum}
\definecolor{burntorange}{HTML}{BF5700}
\definecolor{rubinered}{HTML}{DB487E}
\definecolor{oopsiecolor}{RGB}{203, 39, 191}

\usepackage{tikz}
\usetikzlibrary{shadows}
\usepackage{cleveref}

\title{\LARGE \bf
\ours{}: Learning Transferable VLA Corrections\\ from Semantic Interventions
}
\author{
Arpit Bahety$^{1,*}$ \quad
Skand Peri$^{1,*}$ \quad
Sriniket Ambatipudi$^{1}$ \\[2pt]
Chen Xu$^{2}$ \quad
Haruki Nishimura$^{2}$ \quad
Masha Itkina$^{2}$ \quad
Roberto Mart\'{\i}n-Mart\'{\i}n$^{1}$
}

\begin{document}

\maketitle
\thispagestyle{empty}
\pagestyle{empty}
\begingroup
\renewcommand{\thefootnote}{}
\footnotetext{$^{1}$UT Austin \quad $^{2}$Toyota Research Institute \quad $^{*}$Equal contribution}
\endgroup

%%%%%%%%%%%%%%%%%%%%%%%%%%%%%%%%%%%%%%%%%%%%%%%%%%%%%%%%%%%%%%%%%%%%%%%%%%%%%%%%
\begin{abstract}

Vision-Language-Action (VLA) models can perform a wide range of robotic manipulation tasks, but often fail during deployment.
Teaching them to recover from these failures requires substantial corrective data: 
existing approaches learn to recover from failures by directly associating corrective actions with the observations in which they were demonstrated. For instance, DAgger-like fine-tuning can require tens to hundreds of interactions, yet the resulting policy may still fail to generalize the learned corrections beyond the demonstrated scenarios. We hypothesize that semantics can capture the underlying reasons for failures and their corrections, enabling policies to learn from only a handful of corrective demonstrations and generalize those corrections to novel objects, and scene variations. We propose \ours{}, a framework that uses abduction to infer the intent behind a small set of corrective demonstrations and encodes the inferred knowledge as dense spatial representations and conditional costs. These costs capture when a behavior is undesirable and steer the policy toward more desirable action regions. We experiment across 6 tasks and variations in object poses, scenes, and related tasks, and find that only a handful of corrective demonstrations can enable a VLA to overcome failures across a broad range of novel situations, without requiring new corrective data for each variation. Please refer to our \href{https://robin-lab.cs.utexas.edu/oopsierecover/}{Project Page} for videos.

\end{abstract}

%%%%%%%%%%%%%%%%%%%%%%%%%%%%%%%%%%%%%%%%%%%%%%%%%%%%%%%%%%%%%%%%%%%%%%%%%%%%%%%%

\section{Introduction}
Vision-language-action models (VLAs) provide broad manipulation capabilities~\cite{intelligence2025pi_,bjorck2025gr00t,kim2024openvla}, but can still exhibit systematic failures during deployment~\cite{oopsieverse}, such as collisions, dropping objects, etc.
Human interventions offer a natural source of corrective data~\cite{ross2011reduction, kelly2019hg,liu2025robot, murray2026flowdagger}, yet existing interactive imitation methods generally treat corrections as additional state-action supervision~\cite{ross2011reduction, kelly2019hg}, making generalization depend on collecting enough corrections in similar states.
Humans can instead infer a reusable principle from only a few corrections.
We ask whether a robot can likewise infer what prompted an intervention, how its behavior should be steered to avoid it, and when the same correction should apply again.

An intervention provides only partial evidence about this principle.
Before takeover, the robot executes the VLA actions; afterward, the human provides a preferred continuation.
Instead, the takeover time localizes the context that prompted the intervention and that generalizes to other similar situations, while the subsequent human-controlled segment provides a preferred continuation from the reached state towards a more optimal situation.
A description such as \textit{``keep the mug upright when placing it on the plate''} identifies the semantic property motivating the takeover, which may remain valid despite changes in object(s) poses, robot configuration, background, or even task.
Trajectories alone leave ambiguous which property mattered; language alone lacks the sensorimotor grounding needed for visuomotor control.
Together, they constrain a reusable correction.

\begin{figure}
    \centering
    \includegraphics[width=\linewidth]{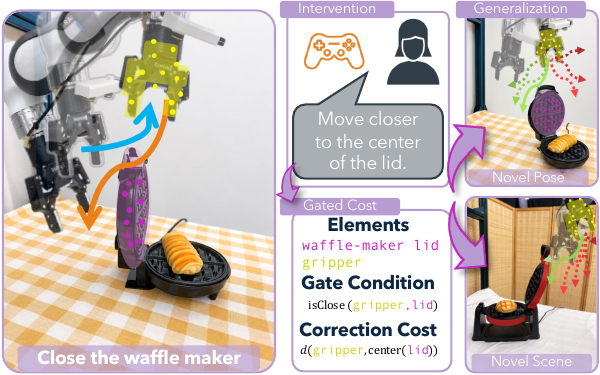}
    \caption{\textbf{Learning reusable corrections from human interventions with \ours{}.} A human corrects a lid-closure failure and describes the intended behavior. \textsc{OopsieRecover} grounds this feedback in semantic point sets, and uses \textit{abduction} -- finding a simple explanation consistent with the observed intervention -- to synthesize a gated correction cost that steers a frozen VLA. (\textit{Right}) The synthesized correction programs generalize: they can be reused under changed object poses and scene layouts, broadening the successful applicability of the VLA.}
    \label{fig:teaser}
    \vspace{-20pt}
\end{figure}

\noindent This ambiguity motivates \ours{}, which formulates correction inference as \textit{abduction}~\cite{kakas1992abductive}: finding a simple explanation consistent with the observed interventions despite missing or imperfect information.
A vision-language model (VLM) leverages its visual-language priors, together with the scene, robot-controlled initial demonstration, human intervention, and language annotation, to propose candidate explanations.
Since several explanations may fit one intervention, \ours{} evaluates each hypothesis across all available interventions and selects explanations that capture the underlying corrective behavior without being overly narrow or permissive.
In this formal-methods inspired procedure, the VLM proposes plausible semantic hypotheses, and the corrective evidence determines whether they are retained or refined.

%Sensorimotor grounding a nd test-time steering. How do we go from costs to executable controL?
We represent the selected explanation as one or more \textit{gated costs}.
Each combines a logical predicate specifying when the correction applies with a continuous cost specifying how the behavior should optimize/change.
E.g., a gate may activate during placement, while its cost penalizes tilting the mug away from upright.
This separation keeps corrections local rather globally biasing the policy.
To execute these costs, \ours{} grounds task-relevant robot parts, object parts, and spatial locations and regions as semantic point sets.
At test time, the VLA samples candidate action chunks, and \ours{} predicts how each candidate would move these point sets, how they evaluate on the gated costs, and which candidate would lead to the lowest cost overall.
The frozen VLA supplies plausible behaviors, while the inferred correction steers its choices whenever the same semantic situation reappears.

%evaluation and contribuitons list
We evaluate \ours{} on 3 tasks in both sim and real using pi0 in RoboCasa and DROID, with only 5-8 corrections per task.
\ours{} improves real-world task-progress from 35.1\% to 61.0\%, a 74\% relative improvement over the next-best baseline, and transfers corrections to unseen poses, objects and scenes. We show that our gated costs generalize to novel variations in object poses and scenes zero-shot, and successfully steer the base policy.
Our contributions are: i) a semantic, abductive formulation of few-shot VLA correction from human-in-the-loop interventions; ii) a generate-and-test procedure that synthesizes and verifies gated costs; and iii) a test-time steering mechanism that applies and generalizes them without retraining the VLA.

\section{Related Works}
\textbf{Interactive Imitation Learning.} Having a human intervene when a robot is about to fail and retraining on the resulting corrective segments is a common approach for improving control policies. DAgger \cite{ross2011reduction} aggregates on-policy states labeled by an expert, while HG-DAgger \cite{kelly2019hg} and IWR \cite{iwr} improve sample efficiency by intervening only when the policy is likely to fail and up-weighting the resulting corrective segments during retraining. SIRIUS \cite{liu2025robot} extends this paradigm to continual learning. Expert Intervention Learning~\cite{spencer2020learning} additionally treats pre-intervention behavior as implicit negative evidence. YAY Robot \cite{shi2024yell} incorporates semantics by training a high-level policy that produces language embeddings for a low-level policy, while OLAF \cite{liu2023interactive} relabels intervened states with actions generated by a VLM conditioned on semantics to improve the base policy, requiring several dozen human corrections~\cite{sun2025dynamic}.

However, these approaches primarily use interventions to relabel the actions and update the policy, limiting the ability of the corrections to generalize beyond the specific situations in which they were collected. In contrast, we focus on extracting a structured representation from the intervention and its associated semantics, allowing the correction to be applied to novel situations in a zero-shot manner.

\textbf{Failure Detection for VLAs.} Detecting failures is a key first step toward recovery. To this end, SAFE~\cite{gu2026safe} trains a multitask failure detector on VLA latent representations to flag anomalous or unsafe executions. FailSafe~\cite{lin2025failsafe} similarly focuses on failure reasoning and recovery signals, identifying failure modes that can trigger intervention. Failure detection is also a central component of \ours{}, however, like most existing methods for failure detection \cite{zhang2026foresight, ren2023robots, sinha2024real, liu2024model, liu2024multi, gu2026safe} these approaches require additional training or fine-tuning, whereas we target the extreme data-scarce regime of learning from only a handful of corrections. We instead leverage point-sets as structured representations and guarded costs to detect when the policy is failing and steer it without requiring any additional training.

\textbf{Inferring objectives/constraints from semantics.} Another line of work is that of inverse reinforcement learning (IRL), where given a (correction induced) demonstration, the goal is to infer the reward function that defines it. One-at-a-time \cite{bajcsy2018learning} extracts the reward by weighting predefined features, and use Bayesian IRL \cite{ramachandran2007bayesian, hadfield2017inverse} to maintain posterior over rewards. Physical-correction methods~\cite{bajcsy2017learning, bobu2020quantifying} similarly infer underlying objectives rather than directly imitating corrected actions. 
Other work uses VLMs to infer rewards~\cite{ma2024eureka,ma2406dreureka,baumli2023vision,pmlr-v235-wang24bn} or constraints from observations~\cite{kumar2026open} or demonstrations~\cite{gumbsch2026demonstrations}. ReKep~\cite{huang2024rekep} and VoxPoser~\cite{huang2023voxposer} further synthesize grounded spatial objectives from VLMs. Programmatic imitation learning (PRL) \cite{tremblay2018synthetically, xin2024programmatic} is another data-efficient approach for representing policies as programs from few demonstrations \cite{patton2024programming}. Code as policies \cite{liang2023code} relaxes the demonstration assumption and entirely generates the policy code using an LLM. While these works represent policies as explicit, human-interpretable programs learned from demonstrations, they typically synthesize a new policy or program \textit{de novo} for each task rather than preserving a base policy. In contrast, \ours{} extracts a local, guarded correction that supplements a frozen, general-purpose VLA only when the relevant semantic context arises, preserving the base policy’s broader capabilities.

\textbf{Steering-based methods.} One line of research focuses on steering generative policies using RL, either by autonomously steering the latent space \cite{wagenmaker2025steering, hong2026tmrl} or through human-in-the-loop interventions \cite{tang2026improving, murray2026flowdagger}. These methods remain relatively sample-inefficient, often requiring several successful trajectories to learn an accurate $Q$-function for effectively steering the base policy. Works like RecoveryChaining~\cite{vats2025recoverychaining} show direct sim2real transfer of recovery behaviors, however, they require re-training of recovery skills for every new scenario. In contrast, we focus on steering policies with as \textit{few} corrections as possible, on the order of only a handful of interventions, while being able to transfer to new task instances zero-shot.

An alternative to RL-based steering is to steer the base policy with a verifier in the loop. FOREWARN~\cite{wu2025foresight} steers the policy by training a latent dynamics model and fine-tuning a VLM to align with it, enabling the VLM to predict the outcomes of imagined trajectories in language and select the most promising one. Do What You Say~\cite{wu2025you} generates multiple candidate trajectories using a simulator and employs a VLM as a verifier to rank them against the VLA's task-level intermediate plan. Most closely related to our work is VLS~\cite{liu2026vls}, which uses a VLM to extract reward functions for each task sub-stage, then generates diverse trajectories through gradient-based guidance and ranks them according to the extracted reward functions before executing the best action sequence. V-GPS~\cite{nakamoto2024steering}, Inference-Time Policy Steering~\cite{wang2025inference}, and DynaGuide~\cite{du2026dynaguide} similarly steer pretrained policies at inference time. Unlike these approaches, \ours{} synthesizes, calibrates, and verifies reusable, context-gated costs from very few human interventions to steer frozen VLAs across tasks and scenes.

\section{\ours{}}
\begin{figure*}[t]
    \centering
    \includegraphics[width=\textwidth]{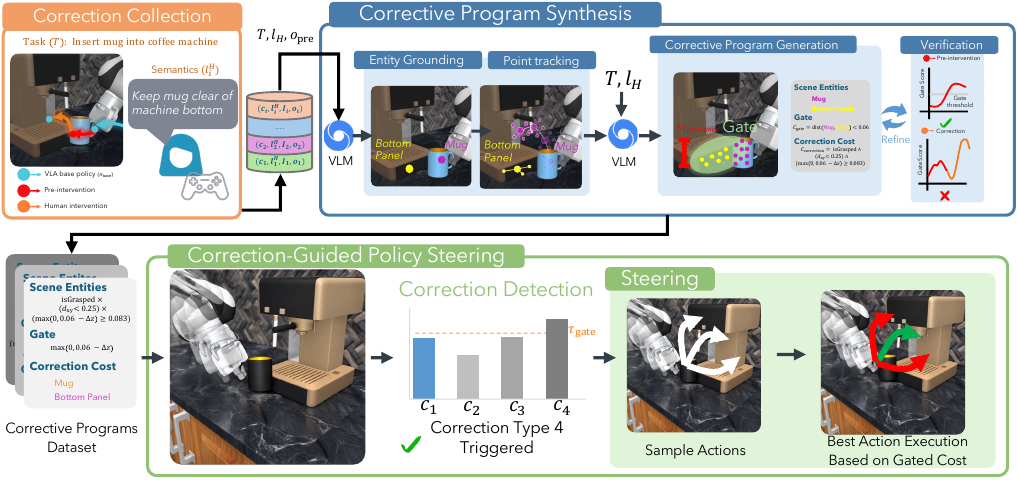}
    \caption{\textbf{\ours{}: From semantic interventions to reusable VLA corrections.} In \ours{}, a human monitors VLA execution and intervenes to provide semantic corrections: a corrective trajectory and a language description for some type of correction, $c_i$, (\textit{\textcolor{orange}{orange}}). We then synthesize the collected set of interventions for each correction type into a \textit{Semantically Grounded Gated Cost} by i) parsing the sensorimotor demonstration as temporal point sets tracked on semantic scene entities, ii) generating gated costs from the motion of the point sets, and iii) verifying these costs by evaluating their capability to explain all human corrections of the same type (\textit{\textcolor{blue}{blue}}).
    During deployment (\textit{\textcolor{green}{green}}), these costs determine when steering is needed and rank candidate actions using the learned corrective programs. This separates the intended correction from its demonstrated trajectory, allowing correction programs to be used in the environment where they are collected and to generalize to new situations, while the base VLA remains frozen.}
    \label{fig:method}

\end{figure*}

\ours{} learns reusable correction objectives from human interventions and uses them to steer a pretrained VLA. \Cref{fig:method} summarizes our method.
A base policy $\pi_{\mathrm{base}}$ executes a task instruction $T$, while a human monitors and intervenes to provide a semantic correction: a corrective trajectory and correction semantics expressed as a free-form language description of the cause of the correction (Sec.~\ref{ss:correction_data}).
\ours{} uses these semantic corrections to synthesize programmatic spatiotemporal objectives over semantically grounded scene entities (Sec.~\ref{ss:correction_synthesis}) in the form of gated correction costs (Sec.~\ref{ss:correction_representation}).
During subsequent execution, \ours{} uses the learned gated correction costs to determine when policy steering is necessary and ranks candidate actions by their predicted costs under a learned dynamics model (Sec.~\ref{ss:correction_steering}).
Further interventions provide additional evidence for refining an existing correction when specified by the human, or introducing a new one.
Throughout this process, the base VLA remains frozen, adaptation occurs through the learned correction objectives; generalization to other domains is thus achieved by using the generalization of the unaltered VLA with the transferrable nature of \ours{}'s gated correction costs.

%%%%%%%%%%%%%%%%%%%%%%%%%%%%%%%%%%%%%%%%%%%%%%%%
\subsection{Collecting Semantic Corrections}
\label{ss:correction_data}

% how do we collect interventions
% We consider a base Vision-Language-Action (VLA) model $\pi_{\mathrm{base}}$ deployed to accomplish a task $\mathcal{T}$. 
During VLA execution, a human can intervene, taking control when the robot's behavior requires correction, providing a corrective action sequence and the correction semantics $\ell_i^H$, and returning control to the VLA when the situation has been corrected~\cite{kelly2019hg}.
For intervention episode $i$, we record all observations and actions as an intervention record $\mathcal{I}_i$, which also includes, when the episode continues after the human returns control, the base-policy window $W^{i}_{\pi_{\text{base}}}$: the $K$ VLA-controlled steps immediately following the correction.
Within an intervention, we distinguish the human-controlled window, $W_{\mathrm{human}}^i$, from a pre-intervention window, $W_{\mathrm{pre}}^i$, the $K$ steps immediately preceding human takeover~\cite{spencer2020learning, liu2023interactive, liu2025robot}.
The pre-intervention window, though part of autonomous execution, provides evidence for the failing behavior that prompted the intervention, while the human-controlled window provides a preferred continuation from the reached state.

After $n$ interventions, the accumulated intervention dataset is $\mathcal D_n =
    \left\{
        (\mathcal I_i,\ell_i^H,c_i, T)
    \right\}_{i=1}^{n}$.
where $c_i$ denotes the human-provided correction type, a semantic category capturing the underlying corrective behavior. For each correction type $c$, we group the corresponding records as
\begin{equation}
    \mathcal D_c =
    \left\{
        (\mathcal I_i,\ell_i^H) :
        (\mathcal I_i,\ell_i^H,c_i)\in\mathcal D_n,\;
        c_i=c
    \right\}.
\end{equation}
\ours{} synthesizes a unifying correction objective from each type-based group of corrective interventions.
To do so, we first define a representation capable of capturing transferable intent.

%%%%%%%%%%%%%%%%%%%%%%%%%%%%%%%%%%%%%%%%%%%%%%%%%%%%%%%
\subsection{Corrections as Semantically Grounded Gated Costs}
\label{ss:correction_representation}

Human corrections can be expressed as spatiotemporal conditions among correction-relevant entities like objects, parts, and regions.
For example, when a mug collides with a coffee machine during placement, the human's corrective action trajectory is specific to the current configuration but the intended correction is more general: maintain clearance from the machine while moving the mug toward the tray. 
Representing this underlying objective captures the transferable intent rather than the specific trajectory used in that single intervention, reducing the need to collect action trajectories in many situations to generalize. We formalize this idea using costs: functions over geometric relations between semantically grounded entities, rather than raw states or actions, enabling transfer across object poses and scenes.
Unlike general-purpose task-objectives in prior works~\cite{ma2024eureka, liu2026vls,wu2025foresight}, our costs capture behavior associated with a particular correction and should only influence execution when that correction is relevant; e.g., keeping the bottom of the mug higher than the base of the coffee machine is only \textit{necessary} near placement.

We therefore represent each correction type as a \textbf{semantically-grounded gated cost} (also termed a \textbf{corrective program}) consisting of entity labels $e$ (e.g., \texttt{``mug''} and \texttt{``bottom panel''}) and three terms: a gate, a correction term, and a task term. 
The gate term, $C^c_{\mathrm{gate}}\in{0,1}$, specifies when the correction is relevant. The correction term, $C^c_{\mathrm{correction}}$, captures the spatiotemporal relations characterizing the human's resolution of the targeted failure, while the task term guides the policy toward actions that preserve task progress, with $\lambda$ controlling this trade-off.

To ground and make such spatiotemporal costs computable from raw observations, we represent the correction-relevant entities as a tuple of entity semantic label and its associated 3D point set: $\mathcal E_{i} =
    (e_i, P^{(j)}_i)$, where
    $P^{(j)} \subset \mathbb R^3$.
An entity can correspond to an object, a selected part, a region, or a single location.
Semantic association specifies which entities participate in the correction, while their point sets provide the geometry needed to ground it and evaluate spatiotemporal relations in the current scene.
Unlike an object label or a single location, a point set retains the observed spatial extent of a region, allowing costs to depend on geometry rather than only a location.
Singleton sets also accommodate individual locations within the same representation, such as the nozzle of the coffee machine.
The total gate cost is computed as:
\begin{equation}
\begin{split}
C^c(\mathcal E)
={}& C^c_{\mathrm{gate}}(\mathcal E)
\Big(
C^c_{\mathrm{correction}}(\mathcal E)
+ \lambda C^c_{\mathrm{task}}(\mathcal E)
\Big).
\end{split}
\label{eq:gated-cost}
\end{equation}

% We then express the correction as an executable cost over these point sets. 
% A relation-based cost can be evaluated using the geometry of the current situation instead of prescribing the coordinates or actions of the original intervention.
% The same cost can also evaluate predicted point sets under candidate actions, providing a direct interface between correction inference and policy steering.
% Transfer therefore builds on grounding the relevant semantic entities in the new situation through point sets and evaluating the learned relation on their geometry through costs.

%%%%%%%%%%%%%%%%%%%%%%%%%%%%%%%%%%%%%%%%%%%%%%%%%%%%%%%%%%%%%%%%%%%%%%%%%%%%
\subsection{Corrective Program Synthesis and Refinement}
\label{ss:correction_synthesis}

For a correction identifier $c$, let $\mathcal D_c$ contain the $n_c\geq 1$ interventions associated with that correction.
Our inference problem is to obtain an executable corrective program that captures their shared intent.
% from the language annotations, the observations, and the recorded trajectories.
% The language annotations suggest the relevant entities and desired behavior, the observations ground those entities in the scene, and the recorded trajectories provide evidence against which a proposed interpretation can be checked.
% cut this if needed
We organize synthesis into four stages: i) grounding the correction-relevant scene entities, ii) proposing a gated cost, iii) calibrating a threshold for the gate score, and iv) verifying and refining that proposal using intervention evidence. 
% The correction identifier determines which program the additional evidence should refine; the language and recorded behavior determine how that program should change.

\subsubsection{Grounding relevant scene entities}
The first step is to identify the entities involved in the correction and ground them with geometry in the recorded observations. 
For \texttt{PlaceMug}, these may include the mug, the obstructing machine region, and the target tray.
We use the task instruction and correction description to localize these entities, then construct and track their 3D point-set representations.

We prompt a VLM with the task language instruction $T$, the correction semantic description $\ell_H$, and the visual observations at the start of the pre-intervention window ($\mathcal{O}_{W_{\mathrm{pre}_0}}$) to identify the correction-relevant scene entities ($e$). For each entity, the VLM specifies a unique seed point $\mathcal{K} \in \mathbb{R}^2$ in the image and determines whether the entity should be represented as a point set or a singleton. For entities requiring a point-set representation, we use the corresponding seed point to prompt a segmentation model~\cite{carion2026sam}, obtaining a 2D region from which we sample pixels and lift them into 3D using depth maps to construct grounded 3D point sets ($\mathcal{P}$). Singleton entites are instead lifted directly from their seed points and represented as individual 3D points.

% Rather than directly predicting segmentation masks, the VLM localizes these elements using a set of seed points
% % \begin{equation}
% %     \mathcal K =
% %     \left\{
% %         k_j
% %     \right\}_{j=1}^{N},
% %     \qquad
% %     k_j \in \mathbb R^2,
% %     \label{eq:grounding_seeds}
% % \end{equation}
% and determines whether each element is represented as a singleton or as point sets. This localization-based formulation provides a simpler interface for the VLM while allowing a dedicated segmentation model~\cite{carion2026sam} to recover the underlying geometry.

% For elements requiring segmentation, we use the corresponding seed points to obtain object or region masks.
% We then sample pixels from each mask and lift them into the robot base frame using the corresponding depth maps and camera intrinsics, yielding the grounded 3D point sets in Eq.~\eqref{eq:semantic_point_sets}.

\subsubsection{Proposing a gated cost}
Given the grounded scene entities, we now want to generate a gated cost. In order to generate a reasonable gate and correction cost, we need the VLM to understand how the correction-relevant entities evolved over time during the pre-intervention and human-correction windows.
To obtain grounded point sets throughout the pre-intervention and human correction trajectories, we track the identified regions across frames and resample pixels before projecting them into 3D at each timestep. 
This avoids computationally expensive and noisy tracking of individual 3D points while preserving correspondence with semantic regions identified by the VLM.

Let $\mathcal{P}_t^{(j)}$ denote the point set corresponding to entity $j$ at timestep $t$. 
Rather than providing the full sequence of point sets to the VLM, we summarize their temporal evolution using all their centroids $\mathbf{x}_t^{(j)}
=
\operatorname{centroid}\!\left(\mathcal{P}_t^{(j)}\right)$ over time $\mathcal{X}_{\mathrm{pre}}^{(j)}
=
\left\{
\mathbf{x}_t^{(j)}
\right\}_{t=t_0}^{t_1}$.
We denote the collection of these trajectories by
$\mathcal{X}_{\mathrm{pre}}$.
They provide a compact representation of the temporal evolution of the relevant scene entities while retaining their semantic correspondence across time.

We prompt the VLM with the task instruction $T$, correction description $\ell_H$, the semantic entity labels $e$ and pre-intervention \& human-correction centroid trajectories $\mathcal{X}_{\mathrm{pre}}$.
The VLM synthesizes a grounded, executable cost $C^c$ with the gate, correction, and task-specific terms in Eq.~\eqref{eq:gated-cost}.
The resulting corrective program is treated as an abductive interpretation of the intervention.

\subsubsection{Intervention-based calibration}
The proposed gated-cost specifies when and what to measure but not how much violation should count as triggering the correction; this threshold must be grounded in the evidence that motivated the intervention itself, rather than left to the VLM's guess. We therefore calibrate the gate against the recorded pre-intervention window before using it both to verify the proposal and, later, to detect when the correction applies and steer the policy at deployment (\cref{ss:correction_steering}).

To measure whether a state exhibits the condition addressed by correction $c$, we combine its gate and correction terms into a gate score:
\begin{equation}
    s_{\mathrm{gate}}^c(\mathcal P)
    =
    C_{\mathrm{gate}}^c(\mathcal P)
    C_{\mathrm{correction}}^c(\mathcal P).
    \label{eq:correction_gate}
\end{equation}
For a recorded trajectory, we write $s_{\mathrm{gate}}^{c,(t)} = s_{\mathrm{gate}}^c(\mathcal P_t)$.

We calibrate a threshold from a recorded pre-intervention window $W_{\mathrm{pre}}$:

\begin{equation}
\begin{aligned}
\tau^{c}_{\text{gate}} &= \min_{i \in \mathcal{D}_c} \; \min_{t \in \mathcal{A}^{c}_{i}} \; s^{c,(t)}_{\text{gate},i}, \\
\mathcal{A}^{c}_{i} &= \left\{\, t \in W^{i}_{\text{pre}} : C^{c}_{\text{gate}}\!\left(\mathcal{P}_t\right) = 1 \,\right\}.
\end{aligned}
\label{eq:gate_calibration}
\end{equation}
This sets the threshold to the lowest gate score observed in the calibration window immediately preceding the human intervention, assuming that reaching this level of cost violation triggered the human intervention.
When synthesizing a gated cost from more than one corrections ($n_c>1$), we take the minimum over episodes \Cref{eq:gate_calibration}, so the detector fires on every recorded intervention.

\subsubsection{Gated cost verification and refinement} 
The initial gated-cost proposal is synthesized before it is checked against the actual intervention actions and may therefore fail to capture the intended correction.
The human's recorded intervention offers a natural test for the original proposal: a correct $C_{\mathrm{gate}}$ and $C_{\mathrm{correction}}^c$ should agree with what the human actually did. 
The task terms $C^c_{\mathrm{task}}$ are excluded from refinement, as our verification checks target the evidence for the human correction.
Given the grounded 3D point sets, we formalize this as two empirical consistency checks:

\noindent \textbf{Ranking check.} First, we check whether the gate score ranks the human correction as an improvement over the pre-intervention behavior. 
We divide the recorded trajectory into equal-length windows for the pre-intervention segment, the human correction, and VLA controlled segments.
For each window $W$, we compute its mean gate score:
\begin{equation}
    \bar{s}^c(W)
    =
    \frac{1}{|W|}
    \sum_{t\in W}
    s_{\mathrm{gate}}^{c,(t)}.
    \label{eq:window_gate_score}
\end{equation}
We require $\bar{s}^c(W_{\mathrm{pre}})
    >
    \bar{s}^c(W_{\mathrm{human}})$

and, when the corresponding base policy's execution window is available, $\bar{s}^c(W_{\mathrm{pre}})
    >
    \bar{s}^c(W_{\pi_\mathrm{base}})$.
These inequalities test whether the synthesized score assigns larger costs to the pre-intervention behavior than to the comparison segments (human correction and $\pi_{\mathrm{base}}$'s execution window). Passing this check indicates that the synthesized cost is consistent with the observed correction direction, assigning lower cost to behaviors that follow or reflect the correction.

\noindent \textbf{Trigger check.} Second, we require the gate score to exceed the calibrated threshold at least once within the pre-intervention segment:
\begin{equation}
    \max_{t\in W_{\mathrm{pre}}}
    s_{\mathrm{gate}}^{c,(t)}
    >
    \tau_{\mathrm{gate}}^c.
    \label{eq:gate_trigger_check}
\end{equation}
This checks whether the calibrated trigger would activate on at least one state preceding the human intervention.
Both checks concern consistency with recorded behavior; they do not guarantee that the program captures the intended correction in every possible situation.

If the verification criteria are satisfied, the gated cost is accepted. 
Otherwise, we refine it using $\left(
        T,\ell_H,\mathcal{X}_{\mathrm{fail}},
        C^c,\mathcal{V}
    \right)$,
where $\mathcal{X}_{\mathrm{fail}}$ is the motion of the point sets in the failed segment, and $\mathcal{V}$ contains the verification evidence, including ranking failures, trigger failures, and gate-calibration failures. 

\noindent \textbf{Refinement.} An unsuccessful verification doesn't mean starting over, it signals which part of the proposal could be wrong. We decompose refinement into two decisions, made sequentially within a single VLM pass: which terms need fixing, and what kind of fix each needs.
First, the VLM determines which cost terms require refinement: $C_{\mathrm{gate}}$, $C_{\mathrm{correction}}$, or both. 
Second, for each selected term, it determines whether to modify its parameters, its logical structure, or both. The VLM then generates a revised gated-cost based on these decisions and the available evidence. 
We repeat verification and refinement until all criteria are satisfied or a maximum refinement budget $B$ is exhausted. If the budget is exhausted without passing both checks, we retain the parameters and logic from the final round.
When multiple corrections share the same type, verification is performed jointly across correction episodes: the ranking and trigger checks are evaluated for each episode and combined using a logical AND ($\land$), such that failure on any episode causes the verification round to fail and triggers further refinement. The gating threshold is calibrated jointly across episodes via \Cref{eq:gate_calibration}, yielding a single threshold that must satisfy the trigger check \Cref{eq:gate_trigger_check} for every correction of that type.

\subsection{Correction-Guided Policy Steering}
\label{ss:correction_steering}
\noindent \textbf{Transferring corrections.} 
At test time, when $\pi_{base}$ encounters a previously seen failure, we cannot simply replay the recorded correction recorded correction --- it was tied to that one scene, and the new one differs. What transfers is not the trajectory but the semantics: the entities the correction referred to, such as a mug or a machine's obstructing edge What transfers is the semantics, not the trajectory.
We therefore task the VLM with re-finding these same correction-relevant entities in the new scene.
To do so, the VLM predicts new seed points conditioned on the reference images with their seed points overlaid, jointly across all correction types. Each transferred seed is then expanded into a 3D point set (or singleton) and tracked as described in \cref{ss:correction_representation}.
For correction type $c$, we compute $s_{\mathrm{gate}}^c(\mathcal P_t)$ and trigger correction-based steering when $s_{\mathrm{gate}}^c(\mathcal P_t)
    >
    \tau_{\mathrm{gate}}^c$.

\noindent \textbf{Policy Steering.} When steering, we first sample $K$ candidate action chunks of horizon $H$: $\{a_{1:H}^{(k)}\}_{k=1}^{K} \sim \pi_{\mathrm{base}}$, and roll each chunk out using a point-based dynamics model.

\begin{equation}
% \begin{split}
    \widehat{\mathcal P}_{1:H}^{(k)}
    =
    f_{\mathrm{dyn}}
    \left(
        \mathcal P_0,a_{1:H}^{(k)}
    \right),
    J^{(k)}
    =
    \sum_{h=1}^{H}
    \gamma^{h-1}
    C_{\mathrm{guard}}^c
    \left(
        \widehat{\mathcal P}_h^{(k)}
    \right).
% \end{split}
\label{eq:predicted_correction_cost}
\end{equation}

Here, $\mathcal P_0$ denotes the current point sets, $f_{\mathrm{dyn}}$ is the dynamics model, and $\gamma \in [0, 1)$ is the discount factor which places greater weight on cost violations occurring earlier in the rollout.

Given these predictions $\widehat{\mathcal P}$, we rank the candidate action chunks according to their predicted gated costs and select the lowest-cost candidate for execution:
\begin{equation}
    k^\star
    =
    \operatorname*{arg\,min}_{k\in\{1,\ldots,K\}}
    J^{(k)},
    \qquad
    a_{1:H}^\star
    =
    a_{1:H}^{(k^\star)}.
    \label{eq:correction_action_selection}
\end{equation}

To increase action diversity, we optionally augment the candidate set with cardinal actions \cite{tang2026improving}. While other approaches can be used to generate diverse candidate chunks \cite{hong2026tmrl, liu2026vls}, we use cardinal actions for their simplicity. Further, if multiple corrections are triggered simultaneously, we rank candidate chunks using the weighted sum of their active steering costs, allowing all triggered corrections to jointly influence action selection. We use the physics from the simulator as $f_{\mathrm{dyn}}$ for our simulation experiments, and a grasp-based dynamics model for real world experiments.

\section{Experiments}
\begin{figure}
    \centering
    \includegraphics[width=\linewidth]{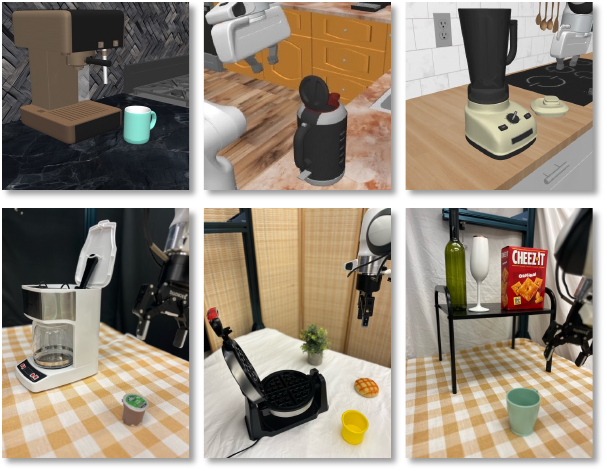}
    \caption{\textbf{Simulation and real-world tasks.} We evaluate a diverse set of tasks that elicit different types of corrections; avoiding collisions, improving placement precision, and correctly articulating objects.}
    \label{fig:tasks}
    \vspace{-2.0em}
\end{figure}

\begin{figure*}[t]
\vspace{-4.0em}
    \centering
    \includegraphics[width=\linewidth]{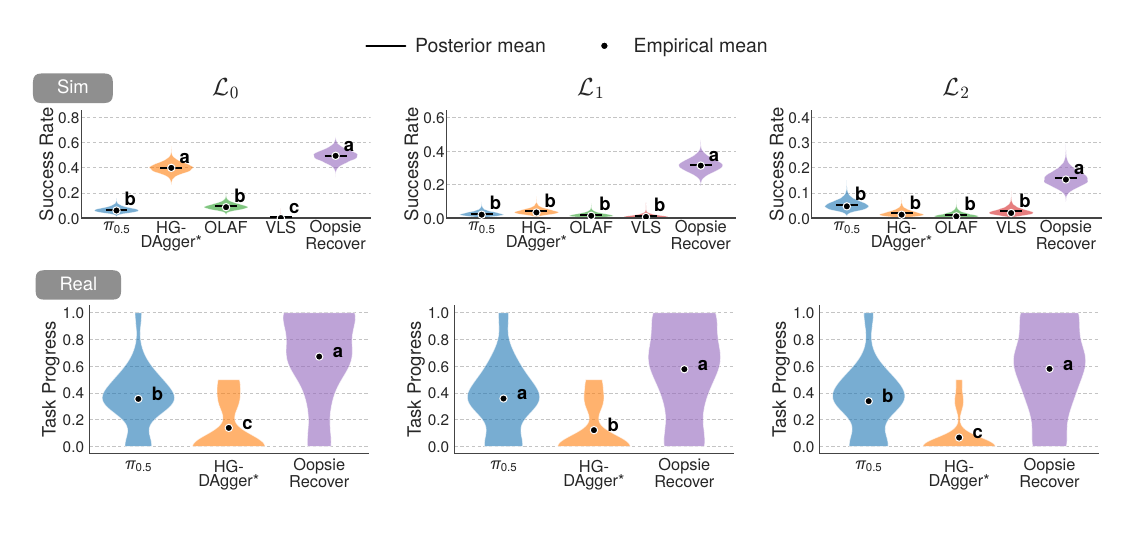}
    \caption{Distribution of success/task-progress across difficulty levels $\mathcal{L}_0$--$\mathcal{L}_2$ (left--right), pooled over three tasks. \textbf{Top (Sim)}: zero-shot success rates; CLD letters (STEP, 95\% confidence) mark statistical separation, shared letters indicate no significant difference; \ours{} leads or ties for best at every level. \textbf{Bottom (Real)}: real-world task-progress scores; \ours~ outperforms the next-best baseline by at least $60\%$ and is statistically separated at $\mathcal{L}_0$ and $\mathcal{L}_2$.}
    \label{fig:q1_sim_real}
\end{figure*}

We aim to answer three questions with our experiments:

\noindent
\textbf{Q1. Can \ours{} learn from a small number of human corrections and generalize to new scenarios?}\\
\noindent
\textit{Simulation Experiments:} We first evaluate the zero-shot generalization of \ours{} in simulation using the RoboCasa benchmark~\cite{robocasa2024}. We consider three tasks: \texttt{CoffeeMugSetup}, \texttt{BlenderLid}, and \texttt{StartKettle}, as illustrated in \Cref{fig:tasks}. For all experiments, we use a $\pi_{0.5}$~\cite{intelligence2025pi_} VLA policy finetuned on the full RoboCasa dataset~\cite{nasiriany2026robocasa365}. For each task, we evaluate all methods under three levels of scene and object-pose variation to assess generalization: $\mathcal{L}_0$ uses the same reset state in which the correction was collected; $\mathcal{L}_1$ uses different initial object poses within the same scene; and $\mathcal{L}_2$ uses different initial object poses in unseen scenes.
We collect $5$–$8$ corrections per task sequentially, one per round, with policy execution in each round steered by all corrections collected in previous rounds. All correction-based baselines are trained on this same data. The baselines are:
\begin{itemize}[leftmargin=*]
    \item \textit{$\pi_{0.5}$}: The base $\pi_0$ policy finetuned on the full RoboCasa dataset, without inference-time steering.
   \item \texttt{HG-DAgger$^*$}: A variant of human-gated DAgger~\cite{kelly2019hg} that uses the same human intervention data collected for \ours{}. We LoRA-finetune $\pi_{0.5}$ on the full set of collected interventions for 20K iterations.
   \item \texttt{OLAF}: A re-implementation of OLAF~\cite{liu2023interactive}, where $\pi_{0.5}$ is finetuned on intervention data in which the pre-intervention actions are relabeled by a VLM.
    \item \texttt{VLS}~\cite{liu2026vls}: An inference-time steering method that extracts a differentiable reward function. Unlike \ours, it relies solely on the task description to generate the reward function, whereas we use human corrections to learn correction programs.
\end{itemize}
%\noindent \textbf{Results.} 

\noindent \textbf{Results.} \Cref{fig:q1_sim_real} (top) shows zero-shot generalization across the three simulation tasks. Statistical significance is determined using the STEP test \cite{snyder2025step}, with compact letter displays~(CLD)~\cite{piepho2004algorithm} denoting significantly different groups at a global confidence level of 95\%. Each method is evaluated over 50 episodes, with 5--8 corrections collected per task. 
The base policy, $\pi_{0.5}$, performs poorly across. Finetuning with the collected interventions (\texttt{HG-DAgger$*$}) improves $\mathcal{L}_0$ success by 40\%, but fails to generalize to $\mathcal{L}_1$ and $\mathcal{L}_2$, where the initial state differs from the correction scenarios. \texttt{OLAF} generally underperforms \texttt{HG-DAgger$*$}, as the raw actions selected by the VLM are often inaccurate. 
Similarly, \texttt{VLS} fails to generalize, as the task rewards provide insufficient signal to effectively steer the policy, resulting in failures such as collisions. These failures highlight where additional correction supervision is needed to learn more informative costs.
In contrast, \ours{} achieves strong performance in $\mathcal{L}_0$ while also generalizing to $\mathcal{L}_1$ and $\mathcal{L}_2$ zero-shot. The CLD analysis confirms this generalization, with \ours{} forming its own statistically separated top-performing group at both difficulty levels. This demonstrates that \ours{} learns transferable corrective behaviors rather than overfitting to the correction scenarios.

\textit{Real World Experiments:} 
We use the DROID hardware setup~\cite{khazatsky2024droid} and the official DROID checkpoint of OpenPI ($\pi_0.5$). We consider three tasks, \texttt{Put Croffle}, \texttt{Load Coffee Pod}, and \texttt{Shelve item} (\Cref{fig:tasks}). Each of these tasks introduces diverse failure modes like misalignment on grasps and placements, being unable to close the lid, toppling fragile objects, etc. In total, we use $3-6$ semantic corrections for each task across all correction types. We again evaluate on the three levels of $\mathcal{L}_0$, $\mathcal{L}_1$ and $\mathcal{L}_2$. Each evaluation is done for 20 rollouts. We consider two baselines: \texttt{$\pi_{0.5}$} and \texttt{HG-DAgger$^*$}. 
We report task progress rather than only binary task success. 

\noindent \Cref{fig:q1_sim_real} (bottom) shows the real-world task-progress results. Since real-world evaluation uses partial task-progress (fraction of stages completed) rather than binary success, statistical significance is instead determined based on the NSCORE sequential test~\cite{snyder2026nscore}, again reported via CLD at a global confidence level of 95\%.
We find that \ours{} outperforms \texttt{HG-DAgger$^*$} by at least $60\%$ on task progress at every difficulty level tested and is statistically separated as the sole best method at $\mathcal{L}_0$ and $\mathcal{L}_2$, tying with $\pi_0.5$ at $\mathcal{L}_1$.

\begin{figure}[t]
    \vspace{-1.0em}
    \centering
    \includegraphics[width=0.7\linewidth]{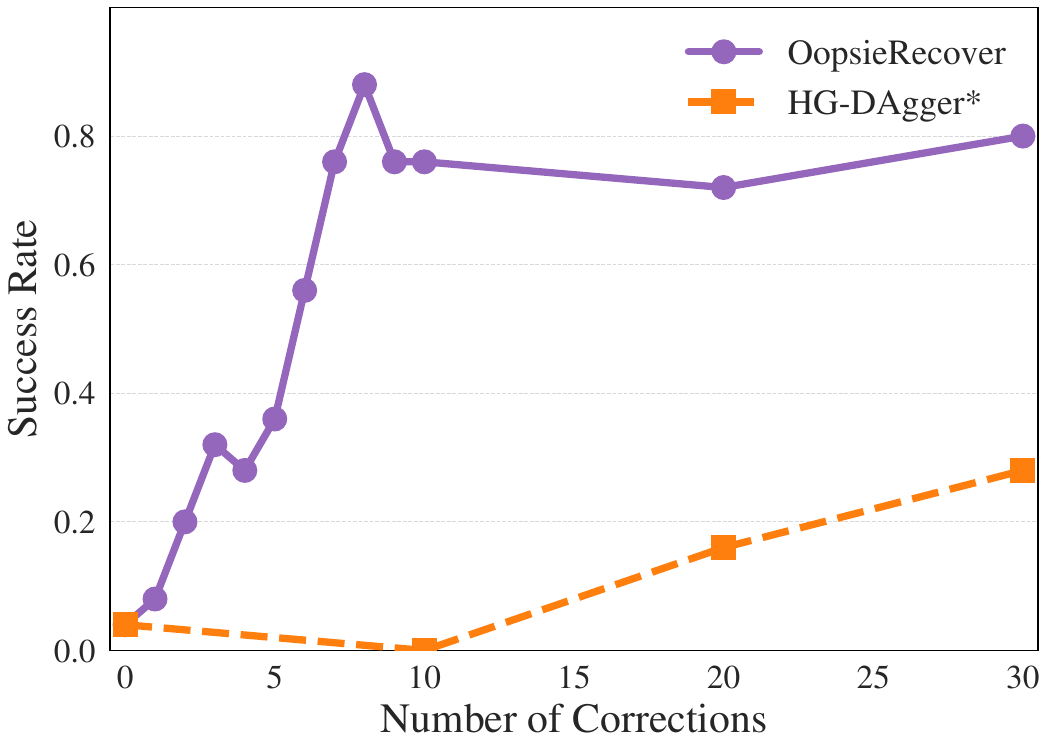}
    \caption{\textbf{Scaling with number of corrections}. On \texttt{CoffeeMug}, \ours{} rapidly reaches an $84\%$ success rate with \textit{as few as 8 corrections}, while \texttt{HG-DAgger$^*$} achieves only $28\%$ despite using three times as many corrections.
}
    \label{fig:q2}
    \vspace{-2.0em}
\end{figure}

\noindent
\textbf{Q2. How does the performance of \ours{} scale with the number of corrections provided?}
We conduct the experiment over multiple rounds, following the standard iterative setup of DAgger on the \texttt{CoffeeMug} task. 
In each round, \ours{} receives \textit{only} one semantically annotated correction, while \texttt{HG-DAgger$^*$} receives ten. We report \ours{} after each round up to round 10, followed by two additional batches of ten corrections, for a total of 20 and 30 corrections. This enables comparison with \texttt{HG-DAgger$^*$} at matched correction counts of 10, 20, and 30.
At every checkpoint, \ours{} updates its behavior by steering with the accumulated corrections, whereas \texttt{HG-DAgger$^*$} is fine-tuned according to the procedure outlined in the previous section.
We collect corrections in randomly sampled scenes from $\mathcal{L}_1$ and evaluate both methods on a held-out test set of 25 scenes with $\mathcal{L}_1$ randomization. We find that \ours{} rapidly improves performance, reaching 84\% success with just 8 corrections before saturating without degradation as additional corrections are added. In contrast, \texttt{HG-DAgger$^*$} improves much more slowly, reaching only 28\% success with the same number of corrections.

\noindent
\textbf{Q3. Can learned correction programs generalize across scenes to detect and prevent the same failure mode?}

To evaluate generalization of learned correction programs, we conduct a controlled experiment using the correction type \textit{``avoid colliding with the steam handle''} from the \texttt{CoffeeMug} task. We learn the corrective program from a single correction episode and test its performance across 10 novel scenes. In each trial, we induce the targeted failure by teleoperating the robot toward the steam handle, and then evaluate whether (a) the correction detector triggers, and (b) whether correction-guided steering prevents the failure over 100 environment steps. In all 10 trials, the correction detector successfully triggers, and correction-guided steering prevents the targeted failure, resulting in a 100\% failure-prevention rate. These results demonstrate that the learned correction program generalizes to novel scenes and enables effective failure prevention beyond the training environment.

% Is the correction doing its job to avoid the failure.
% To evaluate the learned corrective program’s failure detection accuracy, we collect 10 episodes in which a human intervenes to prevent the targeted failure. We then replay these episodes and record when the correction detector, instantiated from the learned program, is triggered. Interestingly, the detector exhibits a conservative response: it triggers before the human intervention in 9 of 10 episodes and only afterward in one episode.

% \item[Q4] How much do verification and refinement contribute to the performance of \ours{}?
% % with and without verification/refinement for CoffeeMug task
% % N corrections, generate cost w and w/o refinement

\section{Limitations}
\ours{} targets corrections expressible as spatial conditions over semantically grounded point sets; its correction representation does not explicitly encode contact forces or unobserved physical properties. 
Its effectiveness depends on reliable scene grounding, point-set tracking, and motion prediction, so perception and dynamics errors can affect both when steering is triggered and which actions are selected.
However, our results with HG-DAgger indicates a promising avenue: distill the corrective behaviors from \ours{} into a image-to-action policy refinement.
Steering selects among a finite set of sampled action chunks and cannot guarantee recovery when this set lacks an effective corrective behavior. 
Finally, the ranking and trigger checks assess consistency with recorded interventions rather than guaranteeing that the inferred program captures the intended correction in every situation.

\section{Conclusion}
We presented \ours{}, a framework that learns transferable corrections for frozen VLAs from a handful of semantically annotated human interventions. It uses abduction to synthesize context-gated costs over semantically grounded 3D point sets, which determine when steering is needed and rank candidate actions by their predicted consequences. Experiments in simulation and the real world show improved execution and transfer across object-pose and scene variations without additional corrective data. Compared to \texttt{HG-DAgger$^*$}, \ours{} achieves higher success with fewer interventions, supporting reusable corrective intent as a data-efficient alternative to directly imitating corrective actions.

\section{Acknowledgments}
This work was conducted in the Robot Interactive Intelligence (RobIn) Lab at The University of Texas at Austin and was supported in part by TRI (R-YFR-04-Martin-Martin), the National Science Foundation (CAREER FAIN 2543745), the NVIDIA Academic Grant Program, Amazon, and DARPA TIAMAT program (HR0011-24-9-0428). The views and conclusions contained herein are those of the authors and should not be interpreted as necessarily representing the official policies or endorsements, either expressed or implied, of the U.S. Government, or any sponsor.

\bibliographystyle{IEEEtran}
\bibliography{references}

\begin{thebibliography}{10}
\providecommand{\url}[1]{#1}
\csname url@samestyle\endcsname
\providecommand{\newblock}{\relax}
\providecommand{\bibinfo}[2]{#2}
\providecommand{\BIBentrySTDinterwordspacing}{\spaceskip=0pt\relax}
\providecommand{\BIBentryALTinterwordstretchfactor}{4}
\providecommand{\BIBentryALTinterwordspacing}{\spaceskip=\fontdimen2\font plus
\BIBentryALTinterwordstretchfactor\fontdimen3\font minus \fontdimen4\font\relax}
\providecommand{\BIBforeignlanguage}[2]{{%
\expandafter\ifx\csname l@#1\endcsname\relax
\typeout{** WARNING: IEEEtran.bst: No hyphenation pattern has been}%
\typeout{** loaded for the language `#1'. Using the pattern for}%
\typeout{** the default language instead.}%
\else
\language=\csname l@#1\endcsname
\fi
#2}}
\providecommand{\BIBdecl}{\relax}
\BIBdecl

\bibitem{intelligence2025pi_}
P.~Intelligence, K.~Black, N.~Brown, J.~Darpinian, K.~Dhabalia, D.~Driess, A.~Esmail, M.~Equi, C.~Finn, N.~Fusai \emph{et~al.}, ``$\pi_{0.5}$: A vision-language-action model with open-world generalization,'' 2025.

\bibitem{bjorck2025gr00t}
J.~Bjorck, F.~Casta{\~n}eda, N.~Cherniadev, X.~Da, R.~Ding, L.~Fan, Y.~Fang, D.~Fox, F.~Hu, S.~Huang \emph{et~al.}, ``Gr00t n1: An open foundation model for generalist humanoid robots,'' 2025.

\bibitem{kim2024openvla}
M.~J. Kim, K.~Pertsch, S.~Karamcheti, T.~Xiao, A.~Balakrishna, S.~Nair, R.~Rafailov, E.~Foster, G.~Lam, P.~Sanketi \emph{et~al.}, ``Openvla: An open-source vision-language-action model,'' 2024.

\bibitem{oopsieverse}
A.~Balaji, A.~Bahety, S.~Ambatipudi, D.~Lam, J.~Xu, and R.~Mart{\'i}n-Mart{\'i}n, ``{OopsieVerse}: A safety benchmark with damage-aware simulation for robot manipulation,'' in \emph{RSS}, 2026.

\bibitem{ross2011reduction}
S.~Ross, G.~Gordon, and D.~Bagnell, ``A reduction of imitation learning and structured prediction to no-regret online learning,'' in \emph{AISTATS}.\hskip 1em plus 0.5em minus 0.4em\relax JMLR, 2011.

\bibitem{kelly2019hg}
M.~Kelly, C.~Sidrane, K.~Driggs-Campbell, and M.~J. Kochenderfer, ``Hg-dagger: Interactive imitation learning with human experts,'' in \emph{(ICRA)}.\hskip 1em plus 0.5em minus 0.4em\relax IEEE, 2019.

\bibitem{liu2025robot}
H.~Liu, S.~Nasiriany, L.~Zhang, Z.~Bao, and Y.~Zhu, ``Robot learning on the job: Human-in-the-loop autonomy and learning during deployment,'' \emph{The International Journal of Robotics Research}, 2025.

\bibitem{murray2026flowdagger}
M.~Murray, D.~Chen, S.~Bagaria, D.~Fortier, T.~Hellebrekers, G.~Mullins, H.~Gajarla, O.~Mees, M.~Cakmak, and A.~Kolobov, ``Flowdagger: Human-in-the-loop adaptation of generative robot policies in latent space,'' 2026.

\bibitem{kakas1992abductive}
A.~C. Kakas, R.~A. Kowalski, and F.~Toni, ``Abductive logic programming,'' \emph{Journal of logic and computation}, 1992.

\bibitem{iwr}
A.~Mandlekar, D.~Xu, R.~Mart{\'\i}n-Mart{\'\i}n, Y.~Zhu, L.~Fei-Fei, and S.~Savarese, ``Human-in-the-loop imitation learning using remote teleoperation,'' 2020.

\bibitem{spencer2020learning}
J.~Spencer, S.~Choudhury, M.~Barnes, M.~Schmittle, M.~Chiang, P.~Ramadge, and S.~Srinivasa, ``Learning from interventions,'' in \emph{Robotics: Science and Systems (RSS)}, 2020.

\bibitem{shi2024yell}
L.~X. Shi, Z.~Hu, T.~Z. Zhao, A.~Sharma, K.~Pertsch, J.~Luo, S.~Levine, and C.~Finn, ``Yell at your robot: Improving on-the-fly from language corrections,'' 2024.

\bibitem{liu2023interactive}
H.~Liu, A.~Chen, Y.~Zhu, A.~Swaminathan, A.~Kolobov, and C.-A. Cheng, ``Interactive robot learning from verbal correction,'' 2023.

\bibitem{sun2025dynamic}
X.~Sun, S.~Yang, Y.~Chen, F.~Fan, Y.~Liang, and D.~Rakita, ``Dynamic rank adjustment in diffusion policies for efficient and flexible training,'' 2025.

\bibitem{gu2026safe}
Q.~Gu, Y.~Ju, S.~Sun, I.~Gilitschenski, H.~Nishimura, M.~Itkina, and F.~Shkurti, ``Safe: Multitask failure detection for vision-language-action models,'' \emph{NeurIPS}, 2026.

\bibitem{lin2025failsafe}
Z.~Lin, J.~Duan, H.~Fang, D.~Fox, R.~Krishna, C.~Tan, and B.~Wen, ``Failsafe: Reasoning and recovery from failures in vision-language-action models,'' 2025.

\bibitem{zhang2026foresight}
H.~Zhang, Y.~Lu, B.~Wang, X.~Kang, Y.-L. Kuo, Z.~Cheng, M.~Wang, and O.~C. Jenkins, ``Foresight: Failure detection for long-horizon robotic manipulation with action-conditioned world model latents,'' 2026.

\bibitem{ren2023robots}
A.~Z. Ren, A.~Dixit, A.~Bodrova, S.~Singh, S.~Tu, N.~Brown, P.~Xu, L.~Takayama, F.~Xia, J.~Varley \emph{et~al.}, ``Robots that ask for help: Uncertainty alignment for large language model planners,'' 2023.

\bibitem{sinha2024real}
R.~Sinha, A.~Elhafsi, C.~Agia, M.~Foutter, E.~Schmerling, and M.~Pavone, ``Real-time anomaly detection and reactive planning with large language models,'' 2024.

\bibitem{liu2024model}
H.~Liu, S.~Dass, R.~Mart{\'\i}n-Mart{\'\i}n, and Y.~Zhu, ``Model-based runtime monitoring with interactive imitation learning,'' in \emph{ICRA}.\hskip 1em plus 0.5em minus 0.4em\relax IEEE, 2024.

\bibitem{liu2024multi}
H.~Liu, Y.~Zhang, E.~Zhang, J.~Liu, C.~Ding, and Y.~Zhu, ``Multi-task interactive robot fleet learning with visual world models,'' 2024.

\bibitem{bajcsy2018learning}
A.~Bajcsy, D.~P. Losey, M.~K. O'Malley, and A.~D. Dragan, ``Learning from physical human corrections, one feature at a time,'' in \emph{HRI}, 2018, pp. 141--149.

\bibitem{ramachandran2007bayesian}
D.~Ramachandran, E.~Amir \emph{et~al.}, ``Bayesian inverse reinforcement learning.'' in \emph{IJCAI}, vol.~7, 2007, pp. 2586--2591.

\bibitem{hadfield2017inverse}
D.~Hadfield-Menell, S.~Milli, P.~Abbeel, S.~J. Russell, and A.~Dragan, ``Inverse reward design,'' \emph{Advances in neural information processing systems}, vol.~30, 2017.

\bibitem{bajcsy2017learning}
A.~Bajcsy, D.~P. Losey, M.~K. O’malley, and A.~D. Dragan, ``Learning robot objectives from physical human interaction,'' in \emph{Conference on robot learning}, 2017.

\bibitem{bobu2020quantifying}
A.~Bobu, A.~Bajcsy, J.~F. Fisac, S.~Deglurkar, and A.~D. Dragan, ``Quantifying hypothesis space misspecification in learning from human-robot demonstrations and physical corrections,'' \emph{arXiv preprint arXiv:2002.00941}, 2020.

\bibitem{ma2024eureka}
Y.~J. Ma, W.~Liang, G.~Wang, D.-A. Huang, O.~Bastani, D.~Jayaraman, Y.~Zhu, J.~Fan \emph{et~al.}, ``Eureka: Human-level reward design via coding large language models,'' in \emph{ICLR}, vol. 2024, 2024, pp. 26\,516--26\,560.

\bibitem{ma2406dreureka}
Y.~J. Ma, W.~Liang, H.~Wang, S.~Wang, Y.~Zhu, L.~Fan, O.~Bastani, and D.~Jayaraman, ``{DrEureka}: Language model guided sim-to-real transfer,'' \emph{RSS}, 2024.

\bibitem{baumli2023vision}
K.~Baumli, S.~Baveja, F.~Behbahani, H.~Chan, G.~Comanici, S.~Flennerhag, M.~Gazeau, K.~Holsheimer, D.~Horgan, M.~Laskin \emph{et~al.}, ``Vision-language models as a source of rewards,'' 2023.

\bibitem{pmlr-v235-wang24bn}
Y.~Wang, Z.~Sun, J.~Zhang, Z.~Xian, E.~Biyik, D.~Held, and Z.~Erickson, ``{RL}-{VLM}-f: Reinforcement learning from vision language foundation model feedback,'' in \emph{ICML}, 2024, pp. 51\,484--51\,501.

\bibitem{kumar2026open}
N.~Kumar, W.~Shen, F.~Ramos, D.~Fox, T.~Lozano-P{\'e}rez, L.~P. Kaelbling, and C.~R. Garrett, ``Open-world task and motion planning via vision-language model generated constraints,'' \emph{IEEE Robotics and Automation Letters}, 2026.

\bibitem{gumbsch2026demonstrations}
C.~Gumbsch, L.~Barcellona, L.~Sch{\"u}nemann, P.~Karageorgis, A.~Zadaianchuk, Z.~Wang, S.~Zakharov, F.~Despinoy, R.~Aljundi, and E.~Gavves, ``From demonstrations to rewards: Test-time prompt optimization for vlm reward models,'' \emph{arXiv preprint arXiv:2606.00083}, 2026.

\bibitem{huang2024rekep}
W.~Huang, C.~Wang, Y.~Li, R.~Zhang, and L.~Fei-Fei, ``Rekep: Spatio-temporal reasoning of relational keypoint constraints for robotic manipulation,'' in \emph{Conference on robot learning}, 2024.

\bibitem{huang2023voxposer}
W.~Huang, C.~Wang, R.~Zhang, Y.~Li, J.~Wu, and L.~Fei-Fei, ``Voxposer: Composable 3d value maps for robotic manipulation with language models,'' \emph{arXiv preprint arXiv:2307.05973}, 2023.

\bibitem{tremblay2018synthetically}
J.~Tremblay, T.~To, A.~Molchanov, S.~Tyree, J.~Kautz, and S.~Birchfield, ``Synthetically trained neural networks for learning human-readable plans from real-world demonstrations,'' in \emph{ICRA}, 2018.

\bibitem{xin2024programmatic}
J.~Xin, L.~Zheng, K.~Rahmani, J.~Wei, J.~Holtz, I.~Dillig, and J.~Biswas, ``Programmatic imitation learning from unlabeled and noisy demonstrations,'' \emph{IEEE RAL}, 2024.

\bibitem{patton2024programming}
N.~Patton, K.~Rahmani, M.~Missula, J.~Biswas, and I.~Dillig, ``Programming-by-demonstration for long-horizon robot tasks,'' \emph{Proceedings of the ACM on Programming Languages}, 2024.

\bibitem{liang2023code}
J.~Liang, W.~Huang, F.~Xia, P.~Xu, K.~Hausman, B.~Ichter, P.~Florence, and A.~Zeng, ``Code as policies: Language model programs for embodied control,'' in \emph{ICRA}.\hskip 1em plus 0.5em minus 0.4em\relax IEEE, 2023, pp. 9493--9500.

\bibitem{wagenmaker2025steering}
A.~Wagenmaker, M.~Nakamoto, Y.~Zhang, S.~Park, W.~Yagoub, A.~Nagabandi, A.~Gupta, and S.~Levine, ``Steering your diffusion policy with latent space reinforcement learning,'' 2025.

\bibitem{hong2026tmrl}
M.~M. Hong, J.~Zhang, A.~Nagabandi, and A.~Gupta, ``Tmrl: Diffusion timestep-modulated pretraining enables exploration for efficient policy finetuning,'' 2026.

\bibitem{tang2026improving}
A.~Tang, W.~Chen, A.~Wagenmaker, C.~Finn, and S.~Levine, ``Improving robotic generalist policies via flow reversal steering,'' 2026.

\bibitem{vats2025recoverychaining}
S.~Vats, D.~K. Jha, M.~Likhachev, O.~Kroemer, and D.~Romeres, ``Recoverychaining: Learning local recovery policies for robust manipulation,'' \emph{IROS}, 2025.

\bibitem{wu2025foresight}
Y.~Wu, R.~Tian, G.~Swamy, and A.~Bajcsy, ``From foresight to forethought: Vlm-in-the-loop policy steering via latent alignment,'' \emph{RSS}, 2025.

\bibitem{wu2025you}
Y.~Wu, A.~Li, T.~Hermans, F.~Ramos, A.~Bajcsy, and C.~P{\~A}{\v{S}}rez-D'Arpino, ``Do what you say: Steering vision-language-action models via runtime reasoning-action alignment verification,'' \emph{ICRA}, 2026.

\bibitem{liu2026vls}
S.~Liu, I.~S. Singh, Y.~Xu, J.~Duan, and R.~Krishna, ``Vls: Steering pretrained robot policies via vision-language models,'' \emph{CoRL}, 2026.

\bibitem{nakamoto2024steering}
M.~Nakamoto, O.~Mees, A.~Kumar, and S.~Levine, ``Steering your generalists: Improving robotic foundation models via value guidance,'' \emph{arXiv preprint arXiv:2410.13816}, 2024.

\bibitem{wang2025inference}
Y.~Wang, L.~Wang, Y.~Du, B.~Sundaralingam, X.~Yang, Y.-W. Chao, C.~P{\'e}rez-D’Arpino, D.~Fox, and J.~Shah, ``Inference-time policy steering through human interactions,'' in \emph{2025 IEEE International Conference on Robotics and Automation (ICRA)}, 2025.

\bibitem{du2026dynaguide}
M.~Du and S.~Song, ``Dynaguide: Steering diffusion polices with active dynamic guidance,'' \emph{Advances in Neural Information Processing Systems}, 2026.

\bibitem{carion2026sam}
N.~Carion, L.~Gustafson, Y.-T. Hu, S.~Debnath, R.~Hu, D.~Suris Coll-Vinent, C.~Ryali, K.~V. Alwala, H.~Khedr, A.~Huang \emph{et~al.}, ``Sam 3: Segment anything with concepts,'' in \emph{ICLR}, vol. 2026, 2026.

\bibitem{robocasa2024}
S.~Nasiriany, A.~Maddukuri, L.~Zhang, A.~Parikh, A.~Lo, A.~Joshi, A.~Mandlekar, and Y.~Zhu, ``Robocasa: Large-scale simulation of everyday tasks for generalist robots,'' in \emph{RSS}, 2024.

\bibitem{nasiriany2026robocasa365}
S.~Nasiriany, S.~Nasiriany, A.~Maddukuri, and Y.~Zhu, ``Robocasa365: A large-scale simulation framework for training and benchmarking generalist robots,'' in \emph{ICLR}, vol. 2026, 2026, pp. 98\,643--98\,667.

\bibitem{snyder2025step}
D.~Snyder, A.~J. Hancock, A.~Badithela, E.~Dixon, P.~Miller, R.~A. Ambrus, A.~Majumdar, M.~Itkina, and H.~Nishimura, ``Is your imitation learning policy better than mine? policy comparison with near-optimal stopping,'' 2025.

\bibitem{piepho2004algorithm}
H.-P. Piepho, ``An algorithm for a letter-based representation of all-pairwise comparisons,'' \emph{Journal of Computational and Graphical Statistics}, vol.~13, no.~2, pp. 456--466, 2004.

\bibitem{khazatsky2024droid}
A.~Khazatsky, K.~Pertsch, S.~Nair, A.~Balakrishna, S.~Dasari, S.~Karamcheti, S.~Nasiriany, M.~K. Srirama, L.~Y. Chen, K.~Ellis, P.~D. Fagan, J.~Hejna, M.~Itkina \emph{et~al.}, ``{DROID}: A large-scale in-the-wild robot manipulation dataset,'' in \emph{RSS}, 2024.

\bibitem{snyder2026nscore}
D.~Snyder, A.~Badithela, N.~Matni, G.~Pappas, A.~Majumdar, M.~Itkina, and H.~Nishimura, ``Beyond binary success: Sample-efficient and statistically rigorous robot policy comparison,'' 2026.

\end{thebibliography}

\end{document}